\pdfoutput=1

\documentclass[11pt]{article}

\usepackage{ACL2023}

\usepackage{times}
\usepackage{latexsym}

\usepackage[T1]{fontenc}

\usepackage[utf8]{inputenc}

\usepackage{microtype}

\usepackage{inconsolata}

\usepackage{url}
\usepackage{latexsym}
\usepackage{amsmath} 
\usepackage{amsfonts}
\usepackage{bm}
\usepackage{graphicx}
\usepackage{booktabs}
\usepackage{siunitx}
\usepackage{hyperref}
\usepackage{subcaption}

\newcounter{xxx}
\title{SPAFIT: Stratified Progressive Adaptation Fine-tuning for Pre-trained Large Language Models}

\author{Samir Arora \\
  Simon Fraser University \\
  Department of Statistics \\
  and Actuarial Science \\
  {\tt samira@sfu.ca} \\\And
  Liangliang Wang \\
  Simon Fraser University \\
  Department of Statistics \\
  and Actuarial Science \\
  {\tt lwa68@sfu.ca} \\}

\begin{document}
\maketitle
\begin{abstract}
  Full fine-tuning is a popular approach to adapt Transformer-based pre-trained large language models to a specific downstream task. However, the substantial requirements for computational power and storage have discouraged its widespread use. Moreover, increasing evidence of catastrophic forgetting and overparameterization in the Transformer architecture has motivated researchers to seek more efficient fine-tuning (PEFT) methods.  Commonly known parameter-efficient fine-tuning methods like LoRA and BitFit are typically applied across all layers of the model. We propose a PEFT method, called Stratified Progressive Adaptation Fine-tuning (SPAFIT), based on the localization of different types of linguistic knowledge to specific layers of the model. Our experiments, conducted on nine tasks from the GLUE benchmark, show that our proposed SPAFIT method outperforms other PEFT methods while fine-tuning only a fraction of the parameters adjusted by other methods. 
\par  
\end{abstract}

\section{Introduction}

\citet{Vaswani-et-al-2017} introduced a new neural network architecture called Transformers, which used concepts of positional encoding and self-attention, and helped many pre-trained language models (PLMs) reach a state of the art quality in various downstream tasks. To use these PLMs for a specific language related application in a specific domain, one needs to perform supervised learning on these models with data specific to that use case. This adaptation of PLMs to one's specific use case is called fine-tuning. If all the parameters in all the layers are allowed to change while adapting the model to this use case, then it is commonly known as full fine-tuning.\par
Despite showing promising results, the use of Transformer-based PLMs coupled with full fine-tuning is constrained by the computational power and memory requirements. This limitation arises from the complex architecture of Transformers, where each layer has millions of parameters accessed during the forward pass. Consequently, the volume of parameters makes the process computationally demanding and imposes challenges in terms of memory and latency during both training and inference phases \cite{Fan-et-al-2020}. 
Empirically, this overparameterization of PLMs is exhibited by \citet{gordon-et-al-2020}, where they found that pruning 30-40\% of parameters from the BERT-base model has no effect on training loss.\par 
In addition to huge computational power, memory, and training data requirements, using full fine-tuning is further discouraged due to its potential of causing catastrophic forgetting.  Catastrophic forgetting happens when a neural network undergoes sequential training on multiple tasks. In this case, the weights essential for the successful execution of task A can be modified to align with the objectives of task B, leading to a significant loss of knowledge related to the initial task \cite{Kirkpatrick-et-al-2017}. 
\par
This potential of catastrophic forgetting is further explored by \citet{kumar-et-al-2022}. Their findings show that when a pre-trained model performs well on dataset A without fine-tuning, and there is a significant difference between dataset A and dataset B, then performing full fine-tuning on dataset A results in poorer accuracy on dataset B compared to the alternative approach of employing linear probing on dataset A. Linear probing, in this context, means tuning the head on dataset A while
freezing all lower layers. These challenges associated with full fine-tuning motivates researchers to develop more parameter efficient fine-tuning (PEFT) methods.\par 
In current literature, PEFT methods typically treat all layers of the model equally, applying fine-tuning techniques uniformly across the board. We propose investigating whether different layers of the model house distinct types of linguistic knowledge. We advocate for adjusting the intensity of fine-tuning methods based on the level of task-specific knowledge within each layer. This approach has the potential to reduce the number of parameters required while preserving comparable performance levels.\par

\section{Literature Review}

In the paradigm of pre-training coupled with fine-tuning, the mechanism behind fine-tuning is not clearly understood, particularly in terms of how features learned during pre-training are transferred to perform well on downstream tasks.
\cite{hu-et-al-2022} underscores the lack of interpretability of these models and emphasizes the current dependence on hypothesis-driven and empirical approaches employed by researchers in the development of the methods. Some of the popular PEFT methods and their possible hypothesis are described below:

\subsection{Adapter based methods}
An adapter itself is a small neural network module which is integrated at multiple locations in PLMs. During fine-tuning, pre-trained parameters of PLMs are frozen and only parameters of these adapters are allowed to change to adapt the model. The number of parameters in these adapter layers are generally very small, often less than one percent of number of parameters in PLMs, thus allowing for parameter efficient fine-tuning. In using this approach, the implied hypothesis is that these small set of parameters residing in adapters, integrated in each layer of the network, can capture the task-specific changes needed in each layer.\par
Two popular Adapter-based methods include the Series Adapter proposed by \citet{Houlsby-et-al-2019} and the Parallel Adapter introduced by \citet{pfeiffer-et-al-2020}. In the Series Adapter, an adapter is added twice within a transformer layer\cite{Vaswani-et-al-2017}. Since a transformer layer comprises two sublayers -- multi-head attention and a feed-forward network, a serial adapter is inserted after each of these sublayers. This placement occurs right after the output of the sublayer is projected back to input size and before the skip connection. Therefore, the output of one adapter 
serves as an input for the next adapter in series. In contrast,
in the Parallel Adapter method, each adapter is a separate module that processes the same input independently. The output of these adapters are then combined before being passed to the next layer.\par  

\subsection{LoRA: Low Rank Adaptation}
In this increasingly popular research area of PEFT methods, \citet{hu-et-al-2022} have made a significant contribution by introducing a novel method called LoRA. The authors propose to represent the necessary adjustments in pre-trained weights to adapt to a specific task through low-rank decomposition matrices, permitting only these matrices to be trainable while keeping pre-trained parameters frozen. By reducing the rank of the matrices containing trainable parameters, LoRA effectively decreases the total number of parameters to be trained. The underlying hypothesis made by the authors is that the adaptation required to fine-tune a PLM for a new task can be effectively represented using a lower-dimensional subspace.\par
If the language model is parameterized over $\Phi$, where $\Phi_0$ represents pre-trained values, $\Phi_{\text{fine-tune}}$ represents values for parameters after fine-tuning, and $\Delta \Phi$ represents the change in weights required to adapt the model to a new task, then $\Phi_{\text{fine-tune}} = \Phi_0 + \Delta \Phi = \Phi_0 + \mathbf{BA}$, where $\Phi_0$, $\Delta \Phi \in \mathbb{R}^{d \times k}$, $\mathbf{B} \in \mathbb{R}^{d \times r}$, and $\mathbf{A} \in \mathbb{R}^{r \times k}$. Here, $r$ is a hyperparameter. \citet{hu-et-al-2022} noted that very small values of $r$ will suffice even for weight matrices from (very) high dimensional space (i.e. high values of $d$ and $k$). Commonly used values for $r$ are in \{2, 4, 8, 16, 64\}. Another hyperparameter in this method is $\alpha$. As noted by the authors, $\Delta \Phi$ is scaled by $\alpha/r$, where $\alpha$ is a constant in $r$. The author recommended to set $\alpha$ to the first $r$ and do not tune it. The additive structure of model also allows for parallelization, which is not possible in Adapter-based methods. Moreover, this approach works against catastrophic forgetting by preserving pre-trained weights and allows to switch between tasks by swapping the LoRA weights. \citet{hu-et-al-2022} limited their experiments to adapting only attention weights in the transformer architecture \cite{Vaswani-et-al-2017}, specifically, $W_q$ (Query weight matrix), $W_k$ (Key weight matrix), $W_v$ (Value weight matrix, and $W_o$ (Output weight matrix).\par

\subsection{BitFit: Bias-terms Fine-tuning}
\citet{Zaken-et-al-2022} proposed a simple yet competitive parameter efficient fine-tuning method. Their approach involves 
freezing a majority of the network and exclusively fine-tuning  the bias terms. The empirical evidence presented by \citet{Zaken-et-al-2022}) shows the important role and impact of bias parameters in significantly altering the network's behavior. The authors advocate further analysis and attention on the bias terms and hypothesise that the changes required to adapt a pre-trained model to a specific task can be accomplished by just allowing bias terms to change while keeping the remainder of the model frozen.
According to \citet{Zaken-et-al-2022}, for small to medium sized training data, BitFit 
exhibits the same or sometimes better accuracy than full fine-tuning. Note, however, that these experiments were conducted only on BERT models. \par 

\section{SPAFIT: Stratified Progressive Adaptation Fine-tuning}

\subsection{Hypothesis and Reasoning}
Among the PEFT methods discussed above and the other ones available in the literature, one particular fine-tuning method is applied across all the layers. One hypothesis which does not get enough spotlight is that earlier layers of the network captures basic linguistic knowledge while the later layers captures more complex task specific knowledge. Therefore, the complexity of fine-tuning should also progress as we go deeper into the network. According to this hypothesis, since basic linguistic knowledge is required in all tasks, some initial layers must remain frozen and need not be fine-tuned, layers in the middle should be trained with somewhat complex fine-tuning methods allowing some number of parameters to change, and layers near the end should be trained with some of the best performing reasonably complex fine-tuning methods allowing reasonable number of parameters to change.\par
Unlike computer vision, where \citet{Zeiler-et-al-2014} used novel visualization techniques to show that deep CNNs trained on image classification dataset learn hierarchy of image features, there is not much work done on associating different layers of PLMs with different types of linguistic knowledge. \citet{peters-et-al-2018} found evidence in support of this hypothesis in bidirectional language models
(biLMs)  and concluded that the  lower layers of biLMs focus on capturing local syntactic relationships. This enables the higher layers to handle longer-range relationships, such as coreference, and to specialize further for the language modeling task at the topmost layers. Another evidence is provided by the empirical study done by \citet{Tenney-et-al-2019} on the BERT model and consistently found that basic syntactic information appears earlier in the network, while high-level semantic information appears at higher layers.\par 

\subsection{Our Model}
Consider a large language model containing $L$ layers of Transformer-based encoders or decoders. We propose to stratify the encoder/decoder layers into three distinct groups. Two hyperparameters are required: $N_1$, indicating the encoder/decoder number that marks the end of the Group 1 and $N_2$, indicating the encoder/decoder number that marks the end of the Group 2. It's important to note that the values of $N_1$ and $N_2$ are determined based on the percentage of the total number of layers the user wishes to allocate to each group (1, 2, and 3). As such, determining the optimal values for these hyperparameters requires experimentation, considering factors such as the architecture of the large language model, the number of training examples available, and the specific downstream task at hand.

Our proposed method, called Stratified Progressive Adaptation Fine-tuning (SPAFIT), is based on the idea that fine-tuning mechanism should become more complex by allowing more parameters to be tuned in group $n+1$ than group $n$. All the parameters in Group 1 are frozen, following the hypothesis that some initial layers captures basic linguistic knowledge and need not be updated. In Group 2, we allow only the bias terms to change in attention and some other sub-layers as found necessary according to the complexity of the task, thus applying BitFit, a simple approach to fine-tuning. In Group 3, we apply LoRA with parameters $r$ and $\alpha$ on some weight matrices of the sub-layers and apply BitFit to other sub-layers of each encoder. \par

\begin{figure*}
    \centering
    \includegraphics[width=0.65\textwidth]{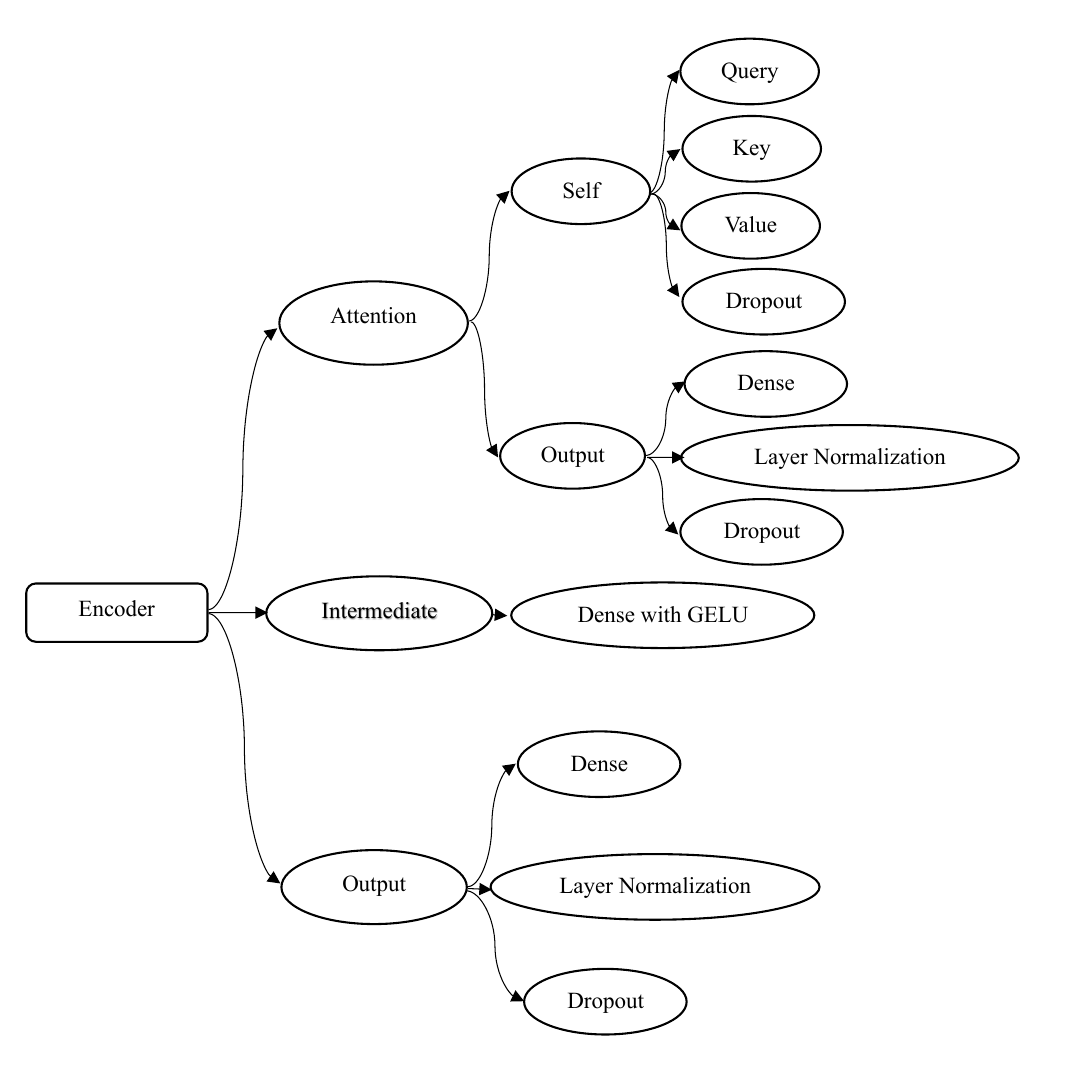}
    \caption{Layer-wise breakdown of an encoder layer of a particular implementation of BERT used for experiments.}
    \label{Figure 1. Layer-wise breakdown of an encoder layer of a particular implementation of BERT used for experiments.}    
\end{figure*}\par

The experiments detailed in this paper utilize the BERT-large-cased model with 24 layers. To comprehend the application of SPAFIT on the BERT-large-cased model, it is crucial to list all the sub-layers within an encoder layer,  according to the specific implementation of the BERT-large-cased model employed in this paper. A comprehensive breakdown of the encoder into its constituent sub-layers is illustrated in Figure \ref{Figure 1. Layer-wise breakdown of an encoder layer of a particular implementation of BERT used for experiments.}. 

The \emph{attention} sub-layer contains \emph{attention.self} sub-layer which applies linear transformations using weight matrices $W_{q}$, $W_{k}$, and $W_{v}$ and bias vectors $b_{q}$, $b_{k}$, and $b_{v}$ to the input to compute query, key, and value matrices $Q(x)$, $K(x)$, and $V(x)$, respectively. These three matrices are used as an input in the multi-head attention computation explained in \citet{Vaswani-et-al-2017}. This gives $H_{1}(x)$ as shown in Equations (\ref{eqn:1}) to (\ref{eqn:4}). 
\begin{align}
    Q(x) &= W_{q}(x) + b_{q} \label{eqn:1}\\  
   K(x) &= W_{k}(x) + b_{k} \\  
   V(x) &= W_{v}(x) + b_{v} \\
   H_{1} =& \text{Multihead\_attention}(Q(x), K(x), V(x)) \label{eqn:4}
\end{align}

Then, the dropout method is employed on $H_{1}(x)$ for regularization, which gives $H_{2}$: 
\begin{align}
   H_{2} &= \text{Dropout}(H_{1}(x))
\end{align}
The \emph{attention.output} sub-layer performs another linear transformation to the output of \emph{attention.self}, $H_{2}$, followed by layer normalization and dropout for regularization as shown in equation (\ref{eqn:6}) and (\ref{eqn:7}).
\begin{align}
    H_{3} &= \text{LayerNorm}(W_{3}(H_{2}) +b_{3})\label{eqn:6} \\
    H_{4} &= \text{Dropout}(H_{3}) \label{eqn:7}
\end{align}
The \emph{intermediate} sub-layer applies a feed-forward neural network as explained by \citet{Vaswani-et-al-2017} along with the GELU activation function on the output of the \emph{attention} sub-layer, $H_{4}$ as shown in equation 8. 
\begin{align}
    H_{5} &= \text{GELU}(W_{5}(H_{4}) + b_{5}) 
\end{align}
Lastly, the \emph{output} sub-layer performs another linear transformation, along with \emph{layer normalization} and \emph{dropout} to transform the output of the \emph{intermediate} sub-layer back to the original dimension as shown in equation (9) and (10).
\begin{align}
    H_{6} &= \text{LayerNorm}(W_{6}(H_{5}) + b_{6}) \\ 
    H_{7} &= \text{Dropout}(H_{6})
\end{align}

\begin{figure*}
    \centering
    \includegraphics[width=0.99\textwidth]{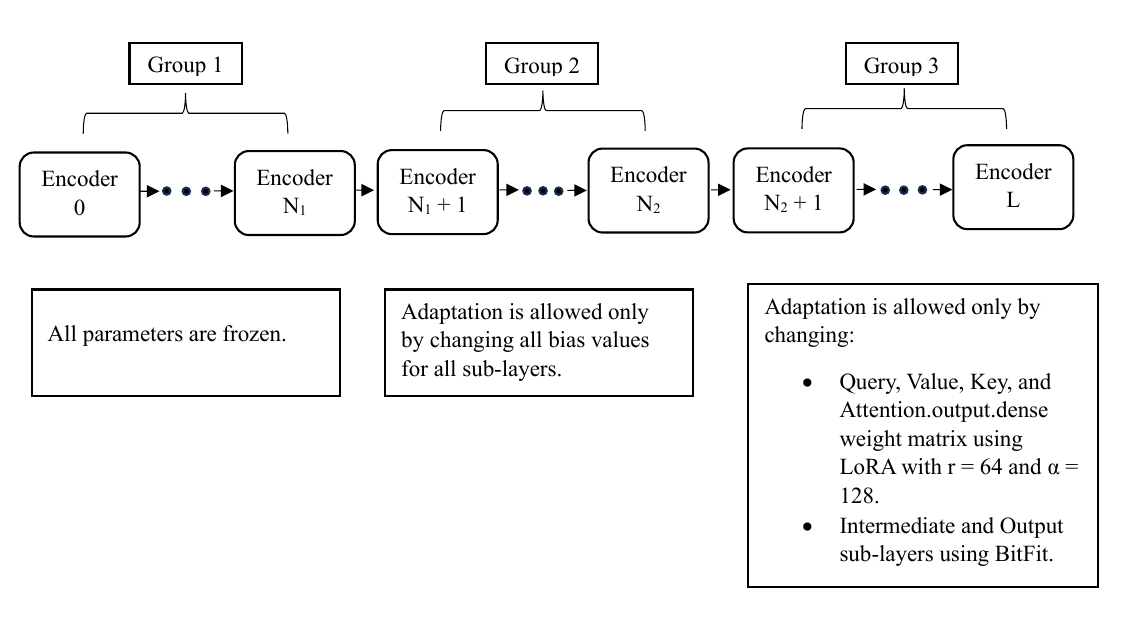}
    \caption{An example implementation of SPAFIT on BERT.}
    \label{Figure 2. An example showing implementation of SPAFIT fine-tuning method on BERT.}    
\end{figure*}

One specific implementation of SPAFIT on the Bert-large-cased model is shown in Figure \ref{Figure 2. An example showing implementation of SPAFIT fine-tuning method on BERT.} . All the parameters in Group 1 remain frozen. In Group 2, adaptation is restricted solely to the modification of the bias terms within all sub-layers of an encoder. For Group 3, weight matrices within the attention sub-layer -- specifically, query, key, value, and attention.output.dense weight matrices -- are allowed to be adapted using LoRA with parameters set to $r$ = 64 and $\alpha$ = 128. Intermediate and output sub-layers are adapted exclusively through the adjustment of bias terms. Decisions regarding the application of BitFit and LoRA to specific layers should be made empirically, dependent on the complexity of the task.\par
\par

\section{Experiments and Results}
The base model used for all the experiments is `BERT-large-cased'. Details about the specific implementation of `BERT-large-cased' used for all the experiments is provided in Figure \ref{Figure 1. Layer-wise breakdown of an encoder layer of a particular implementation of BERT used for experiments.}, presenting a layer-wise breakdown of an encoder layer. Additionally, the implementation of SPAFIT and other fine-tuning methods used in this paper is available in the GitHub repository \footnote{ \url{https://anonymous.4open.science/r/SPAFIT-D326/README.md}}. Please note that this work is licensed under CC BY 4.0 .\par  
In the Full Fine-tuning approach, all parameters are allowed to adjust during adaptation to a new dataset.  While this is a conventional method, it is computationally expensive. The objective of this study is to investigate more cost-effective alternatives that can achieve comparable results. We will assess our proposed SPAFIT models by comparing them  with 
Full Fine-tuning, Full BitFit, and Full LoRA models. \par 
In the Full BitFit model, all the bias parameters in all the layers are allowed to change. In the LoRA models, we set the two hyperparameters $r$ and $\alpha$ to 64 and 128, respectively. We consider two settings of LoRA. In LoRA-I, we only adapt $W_q$, $W_k$, and $W_v$ matrices in the \emph{attention} sub-layer of an encoder; in LoRA-II, besides adapting $W_q$, $W_k$, and $W_v$ matrices in the \emph{attention} sub-layer, we also adapt the dense network in the \emph{output} sub-layer of the attention sub-layer of the encoder. \par In our proposed SPAFIT models, $N_1$ and $N_2$ determine how the $L$ layers of encoders are stratified. None of the parameters in Group 1 will be adapted; in Group 2, only the bias parameters in all sub-layers of each encoder will be adapted. We implement two types of fine-tuning methods for Group 3. FT-I involves applying LoRA-I to the attention sub-layer and BitFit to the intermediate and output sub-layers. In FT-II, we apply LoRA-II to the attention sub-layer and BitFit to the intermediate and output sub-layers.\par
We use the following format to denote a specific SPAFIT model: SPAFIT-[$N_1$]-[$N_2$]-[PEFT Type in Group 3]. For example,  SPAFIT-8-12-II denotes a SPAFIT model where  $N_1=8$, $N_2=12$, and we adapt $W_q$, $W_k$, and $W_v$ matrices and the dense network in the \emph{output} sub-layer of the attention sub-layer along with the bias parameters in intermediate and output sub-layers of each encoder in Group 3.

Table \ref{tab:model_description} provides a summary of all the fine-tuning models used in our experiments.  We use the GLUE dataset \cite{wang-etal-2018-glue} and evaluate the performance of each model on each of the nine tasks in the GLUE dataset. The dataset is detailed in Appendix \ref{section:GLUE Benchmark}.  Training details are provided in Appendix \ref{Section:Training details}.

Table \ref{tab:model_comparison} shows the results of a comparison among various fine-tuning methods. The first column reports the number of parameters that are fine-tuned for each model and the score from the best PEFT method for each task is highlighted in bold.

\begin{table*}
    \small 
    \centering
    \begin{tabular}{lcclll}\toprule
        \textbf{Model} & \textbf{$N_1$} & \textbf{$N_2$} & \textbf{Group 1} & \textbf{Group 2} & \textbf{Group 3}  \\
        \midrule        
        Full Fine-tuning &  & & All parameters & All parameters  & All parameters  \\\midrule    
        Full BitFit & && Bias parameters  & Bias parameters&Bias parameters\\
         & && in all layers &in all layers &in all layers\\ \midrule
        Full LoRA-I  && & LoRA for $W_q$, $W_k$, $W_v$  & LoRA for $W_q$, $W_k$, $W_v$  & LoRA for $W_q$, $W_k$, $W_v$ \\
          && & in \textit{Attention} & in \textit{Attention} & in \textit{Attention}\\\midrule          
        Full LoRA-II  && & LoRA for $W_q$, $W_k$, $W_v$  & LoRA for $W_q$, $W_k$, $W_v$  & LoRA for $W_q$, $W_k$, $W_v$ \\
          && & in \textit{Attention} and & in \textit{Attention} and & in \textit{Attention} and \\
          && & \textit{Attention.Output.Dense} & \textit{Attention.Output.Dense} & \textit{Attention.Output.Dense}\\ \midrule  
          SPAFIT-8-12-I & 8 & 12 & None &  Bias parameters &  FT-I \\
           &  &  &  &  in all sub-layers &    \\      \midrule            
          SPAFIT-8-12-II & 8 & 12 & None &  Bias parameters &  FT-II \\
           &  &  &  &  in all sub-layers &  \\ \midrule                       
           SPAFIT-8-16-II & 8 & 16 & None &  Bias parameters &  FT-II \\
           &  &  &  &  in all sub-layers &  \\    \midrule
            SPAFIT-4-9-I & 4 & 9 & None &  Bias parameters &  FT-I \\
           &  &  &  &  in all sub-layers &  \\  \midrule    
           SPAFIT-4-9-II & 4 & 9 & None &  Bias parameters &  FT-II \\
           &  &  &  &  in all sub-layers &  \\    
          \midrule                    
                     SPAFIT-4-14-II & 4 & 14 & None &  Bias parameters &  FT-II \\
           &  &  &  &  in all sub-layers &  \\  
        \bottomrule
    \end{tabular}
    \caption{Summary of the models and the parameters that are fine-tuned.  In FT-I, LoRA is applied to $W_q$, $W_k$, $W_v$ 
    in \textit{Attention}, and bias parameters are adapted  in the \textit{Intermediate} and \textit{Output} sub-layers;  In FT-II, LoRA is applied to $W_q$, $W_k$, $W_v$ 
    in \textit{Attention} and \textit{Attention.Output.Dense}, and bias parameters are adapted  in the \textit{Intermediate} and \textit{Output} sub-layers.}
    \label{tab:model_description}
\end{table*}

\begin{table*}
    \small 
    \centering
    \begin{tabular}{
        p{2.3cm}| 
        S[table-format=3.2]| 
        *{9}{S[table-format=1.2]} 
    }
        \toprule
        \textbf{Model} & \textbf{Params (M)} & \textbf{CoLA} & \textbf{MNLI} & \textbf{MRPC} & \textbf{QNLI} &  \textbf{QQP} & \textbf{RTE} & \textbf{SST-2} & \textbf{STS-B} & \textbf{WNLI} \\
        \midrule
        Full Fine-tuning & 333.58 & 0.67 & 0.87 & 0.91 & 0.93 & 0.89 & 0.76 & 0.95 & 0.9 & 0.56 \\ \midrule
        Full BitFit & 31.52 & 0.57 & 0.82 & 0.83 & 0.88 & 0.89 & 0.61 & 0.92 & 0.87 & \textbf{0.56} \\ \midrule
        Full LoRA-I & 9.44 & \textbf{0.64} & 0.86 & 0.91 & 0.92 & 0.9 & 0.72 & \textbf{0.93} & 0.88 & \textbf{0.56} \\
        Full LoRA-II & 12.59 & 0.61 & 0.86 & \textbf{0.92} & 0.92 & 0.9 & 0.71 & \textbf{0.93} & 0.9 & \textbf{0.56} \\ \midrule
        SPAFIT-8-12-I  & 4.44 & 0.62 & 0.86 & \textbf{0.92} & 0.92 & 0.9 & 0.73 & \textbf{0.93} & 0.9 & \textbf{0.56} \\
        SPAFIT-8-12-II   & 5.88 & 0.63 & 0.86 & \textbf{0.92} & \textbf{0.93} & 0.9 & 0.74 & \textbf{0.93} & 0.9 & \textbf{0.56} \\
        SPAFIT-8-16-II & 3.81 & 0.62 & 0.86 & 0.91 & \textbf{0.93} & 0.9 & 0.73 & \textbf{0.93} & 0.9 & \textbf{0.56} \\ \midrule
        SPAFIT-4-9-I & 5.65 & \textbf{0.64} & 0.86 & \textbf{0.92} & 0.92 & \textbf{0.91} & 0.74 & \textbf{0.93} & \textbf{0.91} & \textbf{0.56} \\ 
       SPAFIT-4-9-II  & 7.49 & \textbf{0.64} & \textbf{0.87} & \textbf{0.92} & 0.92 & \textbf{0.91} & \textbf{0.76} & \textbf{0.93} & 0.9 & \textbf{0.56} \\
        SPAFIT-4-14-II & 4.89 & 0.63 & 0.86 & \textbf{0.92} & 0.92 & 0.9 & 0.73 & \textbf{0.93} & 0.9 & \textbf{0.56} \\
        \bottomrule
    \end{tabular}
    \caption{Comparison of different fine-tuning methods. Measures reported for CoLA, MRPC, and STS-B are Matthews correlation, F1 score, and Pearson Correlation, respectively. Accuracy score is used as a metric for other tasks. The best performing PEFT method, therefore, not including full fine-tuning is written in bold. Out of all the experiments, the best performance achieved by each fine-tuning method, i.e., max is reported here in this table.}
    \label{tab:model_comparison}
\end{table*}

\section{Discussion}
From Table \ref{tab:model_comparison}, we can observe that, for most tasks, there are PEFT methods capable of achieving performance equal or better than that of  full fine-tuning. Even in tasks where full fine-tuning outperforms, the difference between full fine-tuning and the best-performing PEFT method is very small. This indicates that fine-tuning all parameters in the model may not be necessary to adapt it to a specific downstream task. The finding aligns with the results of 
 \citet{kumar-et-al-2022}, which demonstrate the advantage of avoiding full fine-tuning due to its potential to distort pre-trained features. They show that in cases where two datasets, A and B, significantly differ, a fully fine-tuned pre-trained model on dataset A performs worse on dataset B than linear probing. Between the two extremes of linear probing and full fine-tuning, we have chosen a middle ground based on evidence presented by \citet{tenney-etal-2019-bert}. In our approach, we permit later layers to adapt using increasingly complex PEFT methods while keeping the initial layers frozen.
\par
From Table \ref{tab:model_comparison}, it is evident that BitFit, despite tuning the highest number of parameters (by allowing all bias terms across all layers to adapt to the new dataset while keeping the rest of the layers frozen), is the worst-performing model.  This suggests that some parameters hold more significance than others during fine-tuning. A higher number of tuned parameters does not necessarily translate to better performance. Additional evidence supporting this observation is the performance comparison between SPAFIT-8-12-II and SPAFIT-4-9-I. The former fine-tunes 5.88 million parameters, while the latter fine-tunes 5.65 million parameters. It is notable that the latter achieves the best performance in six out of the nine tasks, whereas the former excels in four of the nine tasks. \par
An interesting finding emerges regarding two tasks where none of the PEFT methods could match or outperform the full fine-tuning performance: CoLA and SST-2. Intriguingly, both CoLA and SST-2 involve tasks with a single sentence as input. In CoLA, the objective is to classify the input sentence into two categories based on its grammatical correctness. In SST-2, the task is to detect the sentiment of the sentence and classify it into two categories: positive or negative.\par
From Table \ref{tab:model_comparison}, we can see that, overall, all SPAFIT models fine-tune significantly fewer parameters than LoRA models and almost all of them perform as well as, or in most cases, better than LoRA models. Therefore, we can conclude that SPAFIT fine-tuning can achieve similar or even better performance than LoRA models while fine-tuning significantly fewer parameters. Two configurations of SPAFIT that perform really well are: SPAFIT-4-9-I and SPAFIT-4-9-II. The second model achieves the best performance in seven out of nine tasks, while the first model excels in six of nine tasks. The only difference between the two models is the application of LoRA on the \emph{attention.output.dense} layer, which increases the number of fine-tuned  parameters by almost two million. The smaller model outperforms the larger model only in one task, STS-B. This difference could be attributed to the smaller training size (approximately 7000 units) of STS-B, which may cause overfitting in the case of the larger model. The smaller model, fine-tuned only 5.65 million parameters (nearly 1.65\% of the total parameters), appears to be a highly efficient model compared to the larger SPAFIT model and certainly in comparison to LoRA and BitFit models.\par
Three tasks where SPAFIT models can outperform full fine-tuning are MRPC, STS-B, and QQP. Interestingly, all these tasks involve sentence similarity. MRPC and QQP are classification tasks, where the goal is to categorize two input sentences into two groups based on whether they are paraphrase of each other. STS-B is slightly different; in this task, the objective is to assign a continuous similarity score.\par

\section{Future Work}
Based on the performance that SPAFIT has shown on the GLUE benchmark, exploring the performance of SPAFIT on complex downstream tasks like summarization could be an  intriguing  extension to this study. Furthermore, exploring a similar stratified approach for models containing both an encoder and a decoder stack would be of interest.\par 
As a long-term goal, we aim to investigate the hypothesis that different types of linguistic knowledge are localized at various layers of a large language model and ascertain its validity.

\section{Limitations}
Despite SPAFIT's commendable performance, it is important to note that all the experiments in this study predominantly involve classification tasks. It is possible that a method limiting the number of parameters to this extent may not perform well on more complex tasks, such as summarization. Another limitation stems from the numerous hyperparameters, including the number of groups, the number of layers in each group, and the complexity variation in fine-tuning changes from group $n$ to $n+1$. This is particularly pertinent as PLMs inherently have many hyperparameters. Furthermore, the implementation of SPAFIT in this  paper still encounters `minor' issues of catastrophic forgetting, given that bias terms are updated during fine-tuning.  However, this challenge can be mitigated by representing changes in the bias vector using a separate vector added to the bias vector.  Lastly, the experiments presented exclusively feature one encoder-based model, Bert-large-cased.  Decoder-based models were not explored  and there is no discussion on extending this fine-tuning methodology to models containing both an encoder and a decoder stack. \par

\section{Ethics Statement}
As advocates for Ethical AI, we would like to emphasize that this research carries the risk of readers assuming it as evidence in favor of the idea that different types of linguistic knowledge are localized in different layers of a large language model. We want to clarify that this is merely a hypothesis behind our efficient fine-tuning method, inspired by CNNs. This work, in no way, confirms the idea of the localization of knowledge in a neural network. 

\bibliography{anthology,custom}
\bibliographystyle{acl_natbib}

\appendix

\section{Appendix}

\subsection{GLUE Benchmark}
\label{section:GLUE Benchmark}
We have used GLUE tasks to evaluate different fine-tuning methods, which are accessed from \href{https://huggingface.co/datasets/glue}{huggingface} website. GLUE is a composite dataset that include the following tasks:  The Corpus of Linguistic Acceptability (CoLA: \cite{warstadt-et-al-2018}) - CC0 1.0 DEED (public domain dedication), The Stanford Sentiment Treebank (SST-2: \cite{socher-et-al-2013} - MIT License (permissive software license), The Microsoft Research Paraphrase Corpus (MRPC: \cite{dolan-et-al-2005} - Microsoft Shared Source License (license providing source code for reference and debugging purposes)), The Quora Question Pairs (QQP: \cite{Iyer-et-al-2017}) - custom (non-commercial) (non-commercial purposes only)), The Semantic Textual Similarity Benchmark (STS-B: \cite{cer-etal-2017} - no license information available), The Multi-Genre Natural Language Inference Corpus (MNLI: 
\cite{bowman-etal-2015-large}
- Creative Commons Share-Alike 3.0 Unported License (allows all content to be freely used, modified, and shared under permissive terms)), The Stanford Question Answering Dataset (QNLI: 
\cite{rajpurkar-etal-2016-squad}
 - CC BY-SA 4.0 (allowing use, remix, and distribute with proper attribution to creators)), The Recognizing Textual Entailment (RTE: \cite{Dagan-et-al-2006} - CC BY 4.0 DEED (permits commercial use, modification, and distribution)), and The Winograd Natural Language Inference (WNLI: \cite{Levesque-et-al-2012} - CC BY 4.0 (allows commercial use, modification and distribution)). Out of all the licenses associated with each of the datasets, the most restrictive is custom (non-commercial). Since we are using this data to train and experiment to find better fine-tuning methods, our use of this artifact is consistent with its terms of use. \par
Basic information on each of these nine tasks is provided below. Please note that the following information comes from the GLUE benchmark website and the dataset page on the \href{https://huggingface.co/datasets/glue}{Hugging Face} website:
\begin{enumerate}
    \item Corpus of Linguistic Acceptability (CoLA): The CoLA dataset contains 10,657 English language sentences from 23 linguistic publications, annotated by their original authors for grammatical acceptability. The training split includes 8.55k examples, the validation set includes 1.04k examples, and the test set includes 1.06k examples. The metric used for this task is Matthew's correlation.
    \item Stanford Sentiment Treebank (SST-2): This dataset is related to a classification task focused on sentiment analysis. The language of this dataset is English too. As per the original paper, the dataset consists of 215,154 unique phrases parsed from  11,855 single sentences extracted from movie reviews, each annotated by three human judges. The training set includes 67.3k rows, the validation set has 872 rows, and the test set has 1.82k rows. The metric used for this task is Accuracy score.
    \item Microsoft Research Paraphrase Corpus (MRPC): The MRPC dataset contains 5,800 pairs of English sentences drawn from news sources on the web. These pairs have been annotated by human judges indicating whether each pair captures a paraphrase/semantic equivalence relationship. One important feature of this dataset is that each training example comes from a unique news article. The training dataset contains 3.67k examples, the validation dataset contains 408 rows, and the test dataset contains 1.73k rows. The metric used for this task is the F1 score. 
    \item Semantic Textual Similarity Benchmark (STS-B): This benchmark dataset consists of 8,628 English sentence pairs from three sources: news, caption, and forum. Out of the total 8,628 sentence pairs: 5,749 pairs are in the training set, 1,500 are in the validation set, and 1,379 are in the test set. These pairs are human-labeled with scores ranging from 0.00 to 5.00. Therefore, the metric used for this dataset is Pearson Correlation.  
    \item Quora Question Pairs (QQP): This dataset contains over 400,000 question pairs in the English language, extracted from the community question-answering website Quora. Each question pair is annotated with a binary value indicating whether the two questions are paraphrases of each other or not. The training dataset contains 364k examples, the validation data contains 40.4k examples, and the test dataset contains 391k rows. The metric used for QQP is accuracy score.
    \item Multi-Genre Natural Language Inference (MNLI): This corpus is a crowd-sourced collection of nearly 433k sentence pairs annotated with textual entailment information. Three labels for each pair are 0 (entailment), 1 (neutral), and 2 (contradiction). Examples that don't have any gold label are marked with a -1 label. Since it is crowd-sourced, it covers a range of genres of spoken and written text and supports a cross-genre generalization evaluation. This corpus also supports only the English language. The dataset is divided into three splits: training (393k examples), validation\_matched (9.82k examples), and validation\_mismatched (9.83k examples). In this work, we are performing fine-tuning only over MNLI Matched. The metric used for this task is the accuracy score.
    \item Stanford Question Answering Dataset (QNLI in GLUE): In this work, we are using SQuAD v1.1. The dataset consists of more than 100,000 question-answer pairs on more than 500 articles in the English language. The authors of the GLUE benchmark convert the task into sentence pair classification by forming a pair between each question and each sentence in the corresponding context and filtering out pairs with low lexical overlap between the question and the context sentence. The task is to determine whether the context sentence contains the answer to the question. The training dataset contains 105k examples, the validation dataset contains 5.46k examples, and the test dataset contains 5.46k examples. The metric used for this task is the accuracy score.
    \item Recognizing Textual Entailment (RTE): This dataset is drawn from a series of annual textual entailment challenges. Examples are constructed based on news and Wikipedia text in the English language. The authors of the GLUE benchmark convert all datasets to a two-class split, whereas for a three-class dataset, they collapse neutral and contradiction into not entailment for consistency. The training split contains 2.49k examples, the validation set contains 277 examples, and the test set contains 3000 observations. The metric used for this task is the accuracy score.
    \item Winograd Natural Language Inference (WNLI): The Winograd Schema Challenge \cite{Levesque-et-al-2012} is a reading comprehension task in which a system must read a sentence with a pronoun and select the referent of the pronoun from a list of choices. The authors of the GLUE benchmark convert this into a sentence pair classification problem by replacing the ambiguous pronoun with each possible referent. The task is to predict if the sentence with the pronoun substituted is entailed by the original sentence. Here, the training set is balanced between the two classes, but the test set is imbalanced with 65\% 'not entailment' examples. The dataset is very small compared to the other eight tasks with 635 observations in the training set, 71 examples in the validation set, and 146 examples in the test set.
\end{enumerate}

\subsection{Training details}
\label{Section:Training details}
Python packages used in this work include: numpy (version 1.23.5), datasets (version 2.16.1), torch (2.1.2), transformers (version 4.37.0.dev0), accelerate (version 0.25.0), bitsandbytes (version 0.41.3.post2), loralib (0.1.2). In addition to that, for implementing LoRA as part of our experiments, we needed git+https://github.com/huggingface/peft.git and git+https://github.com/huggingface/transformers.git. \par
As mentioned earlier, we have only fine-tuned the Bert-large-cased model. We experimented with learning rates in \{2e-3, 6e-3, 2e-5, 6e-5\}. We have found that full fine-tuning and PEFT methods achieved their best performance with learning rates of 2e-5 and 6e-5, respectively. These best performances are the ones reported in Table \ref{tab:model_comparison}. The batch size used for all tasks except MNLI, QNLI, and QQP tasks in the GLUE benchmark is 16. For MNLI, QNLI, and QQP tasks, the batch size
used is 8, as a workaround against the Out Of Memory (OOM) error. 
The optimization algorithm used for fine-tuning is AdamW with a weight decay of 0.01 across all tasks and all fine-tuning methods. The number of epochs used is 10 and remains the same across all tasks and all fine-tuning methods. \par
The experiments are performed using the Tesla V100 GPU available in the Google Collab notebook. As part of the computational budget, please note that we have used around 1500-1800 compute units in Google Collab notebooks during these experiments. 

\end{document}